\documentclass{article}

\makeatletter
\def\input@path{{styles/}}
\makeatother
\usepackage[preprint]{colm2026_conference}
\usepackage{fontspec}

\normalfont
\usepackage{microtype}
\usepackage{graphicx}
\usepackage{trimclip}
\usepackage{xcolor}
\usepackage{booktabs}
\usepackage{array}
\usepackage{colortbl}
\usepackage{float}
\usepackage{tikz}
\usepackage{tcolorbox}
\usepackage{pgfplots}
\usepackage{amsmath}
\usepackage{amssymb}
\pgfplotsset{compat=1.18}
\usepgfplotslibrary{groupplots}
\usepackage{hyperref}
\usepackage{url}
\definecolor{abyss}{HTML}{121D36}
\definecolor{polarnight}{HTML}{1A2947}
\definecolor{nebula}{HTML}{2B3F66}
\definecolor{steeltrail}{HTML}{6D87BD}
\definecolor{skytrail}{HTML}{8FA8D8}
\definecolor{starlight}{HTML}{DFE7F5}
\definecolor{warmstar}{HTML}{E8D9C4}
\definecolor{allsparkwordmark}{HTML}{16233F}
\definecolor{allsparkspark}{HTML}{4A659C}
\definecolor{electricblue}{HTML}{3866FF}
\definecolor{covercream}{HTML}{EEF3FA}
\definecolor{coveraccent}{HTML}{3866FF}
\colorlet{pevekpurple}{skytrail}
\colorlet{bargray}{steeltrail}
\colorlet{barlgray}{starlight}

\newfontfamily\outfit[
  Path=assets/fonts/,
  UprightFont=Outfit-Regular.ttf,
  BoldFont=Outfit-SemiBold.ttf
]{Outfit}

\hypersetup{
  colorlinks=true,
  linkcolor=electricblue,
  citecolor=electricblue,
  urlcolor=coveraccent,
  filecolor=electricblue
}
\setcitestyle{numbers,square,comma,sort&compress}

\newcommand{\reporttitle}{Climbing to the Search Frontier}
\title{\reporttitle}
\author{AllSpark Team}

\begin{document}

\fancyhead{}
\renewcommand{\headrulewidth}{0pt}
\color{abyss}
\thispagestyle{empty}

\vspace*{-0.44in}
\begin{tcolorbox}[
  width=\linewidth,
  colback=covercream,
  colframe=covercream,
  boxrule=0pt,
  arc=14pt,
  outer arc=14pt,
  boxsep=0pt,
  left=20pt,
  right=20pt,
  top=13pt,
  bottom=11pt
]
  {\outfit\fontsize{21.5}{25.5}\selectfont\bfseries\centering
    \textcolor{coveraccent}{Iris:}\hspace{0.25em}\reporttitle\par}
  \vspace{1.45em}
  {\bfseries\centering AllSpark Team\par}

  \vspace{0.55em}
  {\centering\fontsize{9}{11}\selectfont\itshape
   \textcolor{steeltrail}{Do the right things, and do things right.}\par}
  \vspace{0.55em}
  \begingroup
  \normalfont
  \setlength{\parindent}{0pt}
  \setlength{\parskip}{0pt}

We present {Iris-mini and Iris-pro}, two search agents trained at the 35B-A3B and 397B-A17B scales, together with the data pipeline and training recipe behind them. Tasks are reverse-constructed from the hyperlink structure of a web corpus: we author multi-hop chains over an entity graph distilled from a seed page and its out-links, rewrite every non-answer entity into a descriptive reference so that no clue can be resolved by string matching, and admit only questions that a reference model fails closed-book yet solves once the supporting evidence is supplied. These questions are then turned into trajectories, which are filtered twice before supervised fine-tuning (SFT): {first at the trajectory level for correctness, degeneracy, and search depth, and then at the turn level by a judge whose rubric is induced from the data rather than written by hand}. The policy is then optimized by reinforcement learning (RL) against live search, with the reward judge and the observation summarizer served inside the training cluster, and with over-long rollouts interrupted at the request level and resumed from their committed prefix at the next step. We alternate the two stages in a procedure we call \emph{SFT--RL climbing}, returning the hardest solved and most efficient rollouts of each RL round to the next supervised pass. Because inference-time context management is worth more on these benchmarks than most reported differences between systems, we evaluate every benchmark both with and without it, holding the tool set, the context limit, and the judge fixed. All results come from a single ReAct agent, with no sub-agents and no test-time verification. With management enabled, on BrowseComp, BrowseComp-ZH, DeepSearchQA, and Humanity's Last Exam the two models reach $82.2/84.8/86.9/52.3$ and $88.6/85.1/92.9/56.4$, the strongest overall results among open-source search agents in their respective parameter ranges. We plan to release the model weights together with the complete recipe for data construction, training, and evaluation.
  \par
  \endgroup

  \vspace{0.65em}
  \noindent
  \begin{minipage}[b]{0.63\linewidth}
    \outfit\fontsize{8.4}{10.2}\selectfont
    \textbf{Date:} September 4, 2026\\[-0.1em]
    \textbf{Resources:}
    \href{https://github.com/AllSpark-Research/Iris}{GitHub}\,\textperiodcentered\,
    \href{https://huggingface.co/collections/AllSpark-Research/iris}{HuggingFace}\
  \end{minipage}%
  \hfill
  \begin{minipage}[b]{0.33\linewidth}
    \raggedleft
    \raisebox{-0.30em}{\includegraphics[height=16pt]{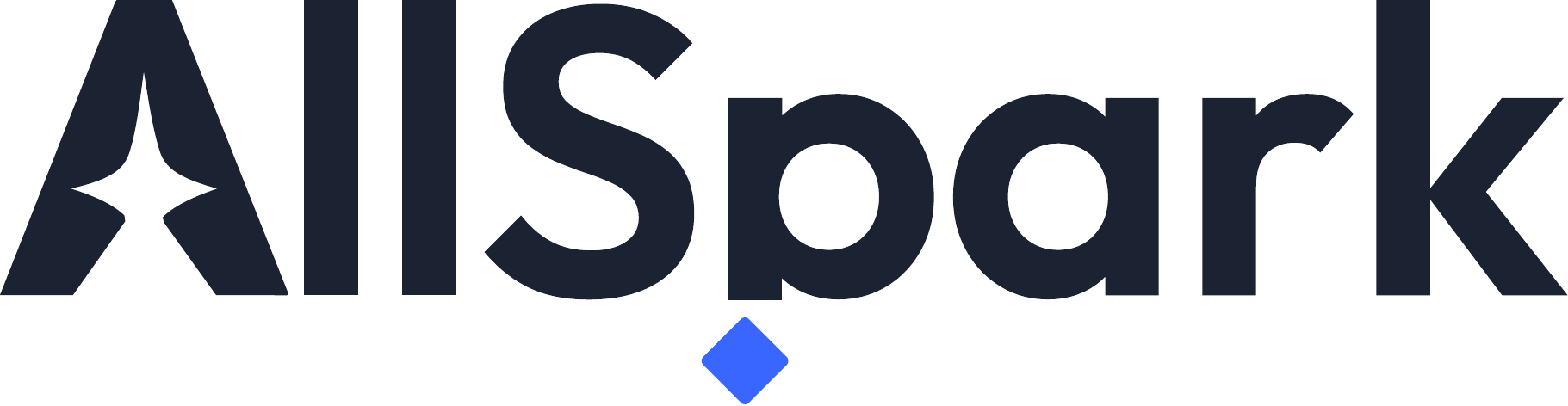}}%
  \end{minipage}
\end{tcolorbox}


\vspace{3em}
\begin{figure}[H]
    \centering

    \begin{minipage}[t]{0.495\linewidth}
        \centering
        \includegraphics[width=\linewidth]{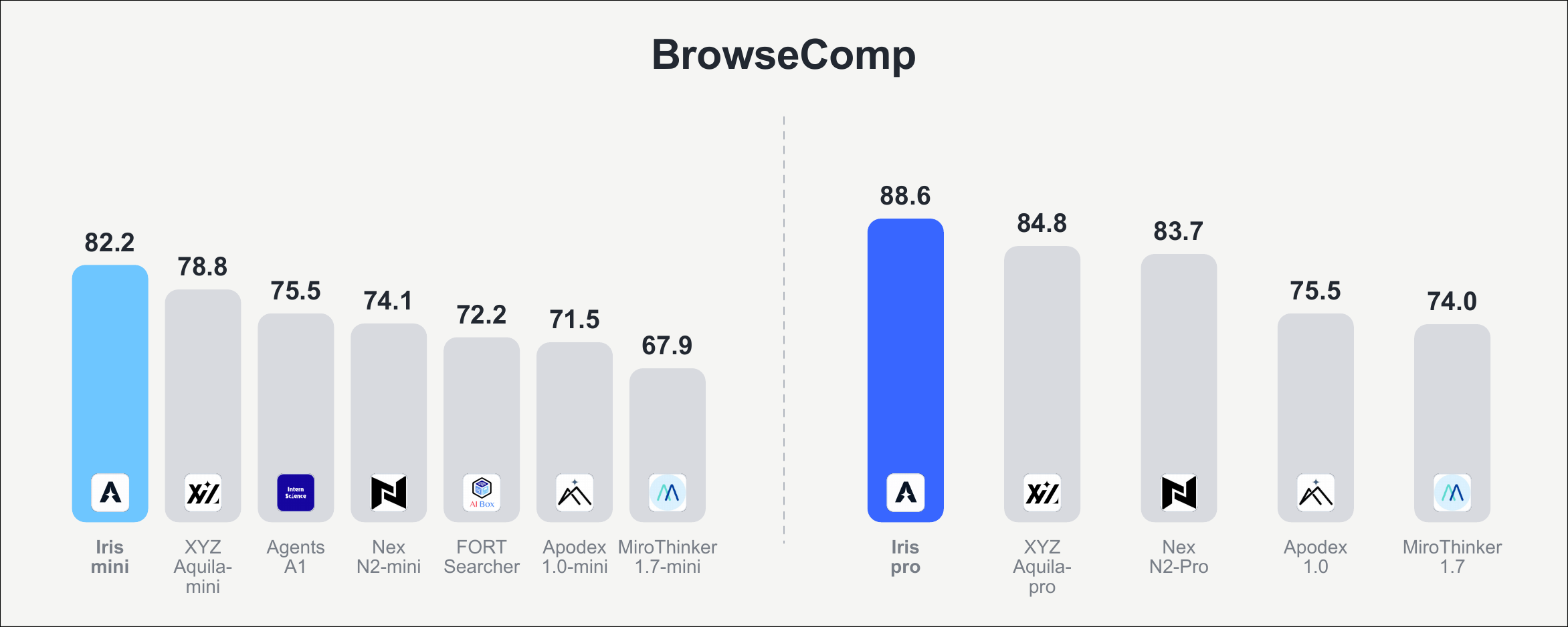}
    \end{minipage}%
    \hfill
    \begin{minipage}[t]{0.495\linewidth}
        \centering
        \includegraphics[width=\linewidth]{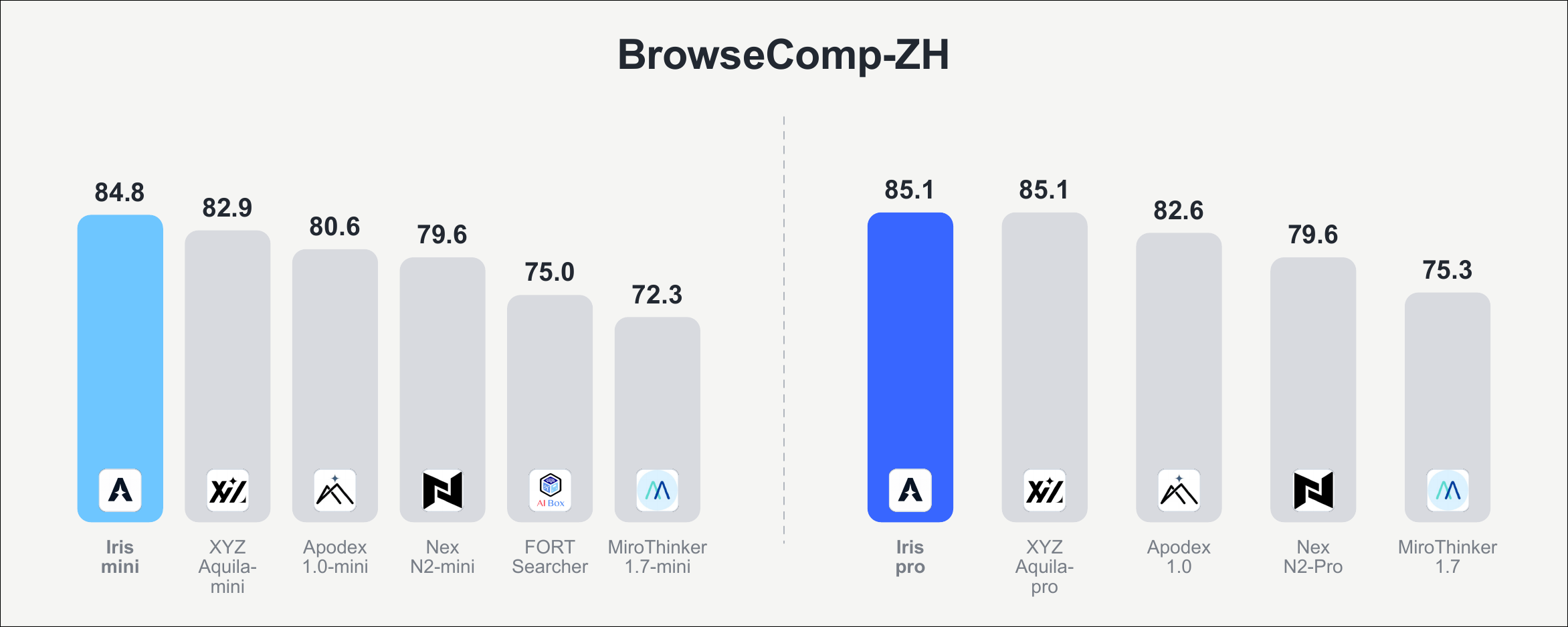}
    \end{minipage}

    \vspace{0.35em}

    \begin{minipage}[t]{0.495\linewidth}
        \centering
        \includegraphics[width=\linewidth]{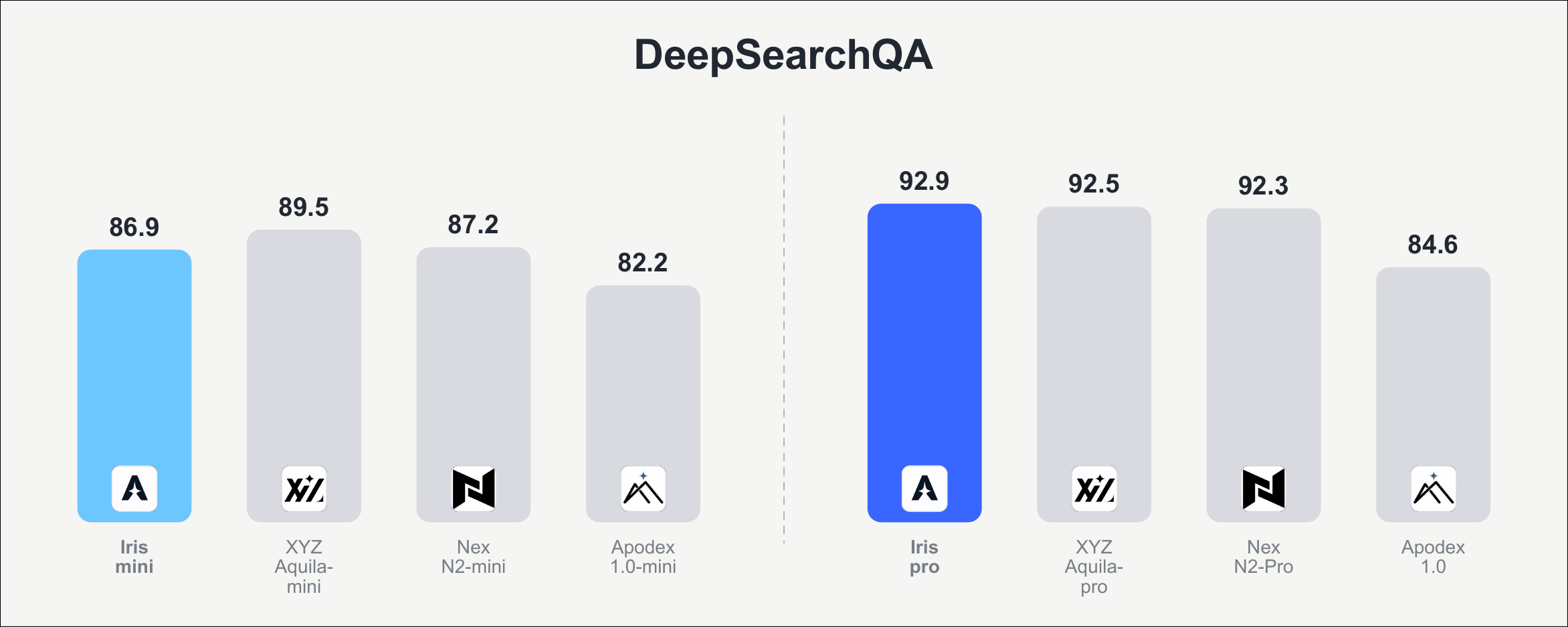}
    \end{minipage}%
    \hfill
    \begin{minipage}[t]{0.495\linewidth}
        \centering
        \includegraphics[width=\linewidth]{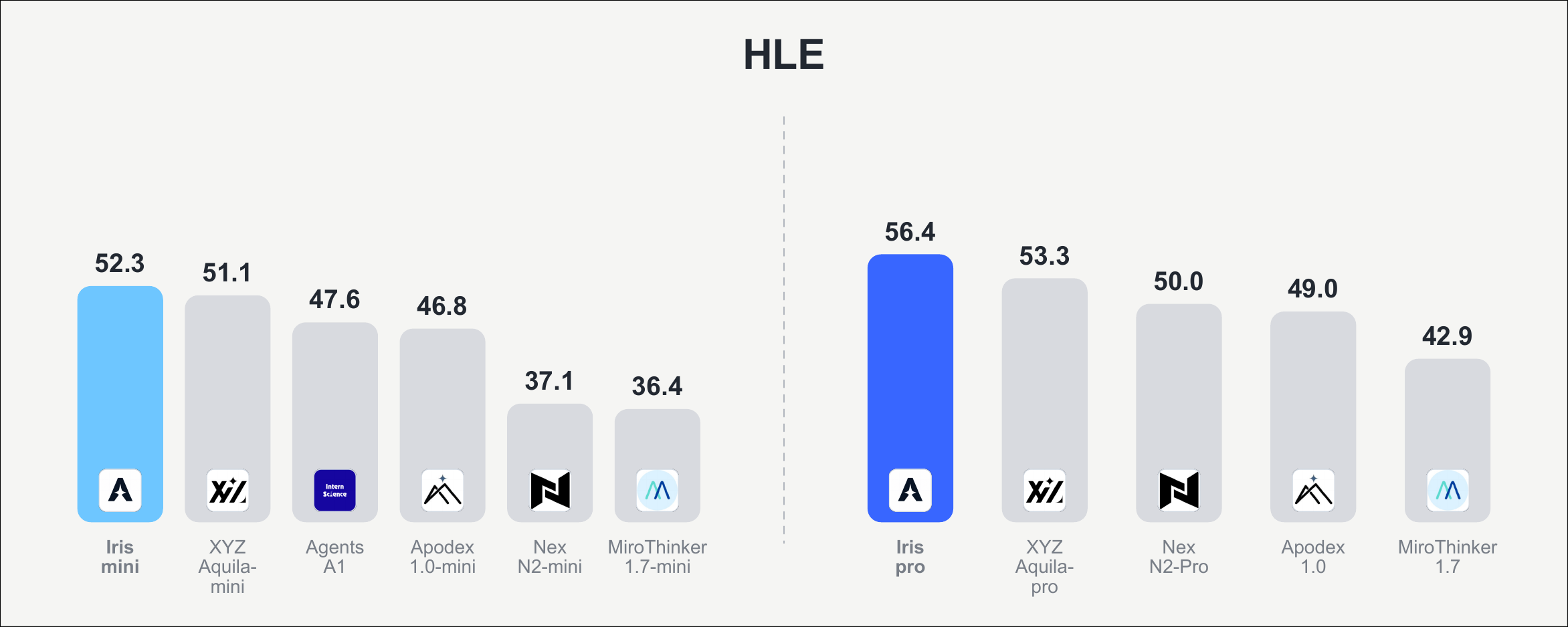}
    \end{minipage}

    \vspace{-0.2em}

    \caption{
    Performance comparison across four agentic search benchmarks.
    }
    \label{fig:benchmark-overview}
\end{figure}
\clearpage

\section{Introduction}
\label{sec:intro}

Search agents extend language models beyond closed-form reasoning by allowing them to interact with external tools and retrieve information from dynamic environments. A capable search agent must not only reason about the question, but also decide what to search, how to interpret retrieved evidence, when to continue exploring, and when the available evidence is sufficient to answer. This makes search a fundamentally different setting from conventional language-model evaluation, where the task, context, and computation budget are largely fixed in advance. Building reliable search agents therefore requires not only stronger models, but also effective training data, long-horizon interaction strategies, and evaluation protocols.

Recent work has made rapid progress toward this goal. Early systems introduced tool use and interleaved reasoning and acting through supervised or self-supervised training~\cite{webgpt, toolformer, react}, followed by reinforcement learning with retrieval from static corpora and, more recently, the live web~\cite{searcho1, searchr1, r1searcher, deepresearcher}. Because naturally occurring web questions are often too easy for training, several studies construct harder tasks by traversing hyperlink or knowledge graphs and masking entities along the resulting paths~\cite{webdancer, websailor, webshaper, deepdive}. Recent systems have further combined synthetic question generation, trajectory supervision, reinforcement learning, and long-context inference into increasingly standardized pipelines~\cite{tongyidr, openseeker, mirothinker, redsearcher, fortsearcher, kimik3, deepseek, apodex, xyz, agentsa1, nex}. Despite these advances, substantially different performance can still arise from differences in the inference-time harness rather than from differences in the underlying policy.

A particularly important factor is \emph{context management} {(CM)}. Long-horizon search can exhaust the available context before the agent has resolved all required constraints, making the effective search budget much smaller than the nominal context window. Existing systems address this through trajectory summarization, state compression, selective history removal, or full context reset~\cite{resum, agentfold, deepseek}. The impact can be substantial when search trajectories are long, while the benefit diminishes as the available context grows or the agent requires fewer interaction steps~\cite{kimik3, fortsearcher, redsearcher}. This suggests that {CM} should be viewed not simply as an implementation detail, but as part of the effective inference system. Reporting only results under a managed context can therefore obscure how much performance comes from the policy itself and how much comes from the surrounding harness.


In this report, we present an end-to-end recipe for {the data construction, training, and evaluation of} strong search agents. Our training pipeline starts from automatically constructed multi-hop tasks derived from web-graph structure. We remove easily searchable anchors through entity rewriting and retain only questions that are both sufficiently difficult and objectively verifiable (Section~\ref{sec:data}). 
We then collect trajectories from a strong teacher and apply multi-stage filtering at both the trajectory and turn levels to improve the quality of supervised training data. 
The resulting policy is optimized with reinforcement learning {against} live search, using an in-cluster judge and observation summarizer, and SFT and RL are alternated iteratively so that successful behaviors discovered during each rollout stage are reinforced before the next round of exploration (Section~\ref{sec:training}).

We evaluate the resulting agents under two inference regimes: with and without {CM}. All other major components, including the tool interface, context budget, and judging procedure, are held fixed to make the effect of {CM} directly measurable. 
As shown in {Figure}~\ref{fig:benchmark-overview}, our Iris-mini and Iris-pro systems achieve the best overall performance among open-source search agents across four challenging benchmarks: BrowseComp~\cite{browsecomp}, BrowseComp-ZH~\cite{browsecompzh}, DeepSearchQA~\cite{deepsearchqa}, and Humanity's Last Exam (HLE)~\cite{hle}.

Our main contributions are summarized as follows:
\begin{itemize}
    \item We develop an end-to-end pipeline for constructing and verifying challenging multi-hop search tasks from web structure, together with trajectory-level and turn-level filtering to obtain high-quality training data.

    \item We combine filtered SFT, RL with live web search, and iterative SFT--RL climbing, allowing high-quality trajectories discovered during RL to be fed back into subsequent supervised training rounds.

    \item We plan to release the model weights, and key components of the data construction, training, and evaluation pipelines to facilitate reproduction and further research on search agents.
\end{itemize}


\section{Data Pipeline}
\label{sec:data}

A capable search agent can be trained on questions that (i) cannot be answered from parametric memory alone and (ii) require composing evidence dispersed across several sources. Naturally occurring web questions rarely satisfy both conditions, and hand-written questions are expensive and hard to scale. We therefore build a fully LLM-driven data pipeline that reverse-constructs questions from the hyperlink structure of a web corpus. 
The pipeline is organized into three stages: web-graph construction, {task synthesis}, and dual-criteria verification.

\subsection{Web-Graph Construction}
\label{sec:qa-graph}

We model the corpus as a directed graph whose nodes are pages and whose edges are hyperlinks,
\begin{equation}
G=(V,E), \qquad \mathrm{Out}(v)=\{\,u\in V \mid (v,u)\in E\,\}.
\label{eq:graph}
\end{equation}

Each synthesis instance begins from a seed page drawn by a sampling policy,
\begin{equation}
v_{0}\sim P_{\text{seed}}(V),
\label{eq:seed}
\end{equation}
for which we use the answer-anchored mode that fixes a target answer entity and retrieves pages describing it. We then expand the seed along its out-links into a local subgraph
\begin{equation}
G_{\text{sub}}=\bigl(\{v_{0}\}\cup N,\;E_{\text{sub}}\bigr),\qquad
N=\{v_{i}\}_{i=1}^{k}\subseteq \mathrm{Out}(v_{0}),
\label{eq:subgraph}
\end{equation}
retaining the full text of each page truncated to a fixed budget. Because the reader services that render pages strip inline anchors, we recover the true out-link set of $v_{0}$ from a structured semantic mirror of the corpus (RDF triples) together with rendered page markup, and merge the two sources to maximize link recall.

\subsection{Task Synthesis}
\label{sec:qa-synth}

Given a subgraph $G_{\text{sub}}$, this stage produces a single obscured multi-hop question through three steps: distilling the pages into an entity graph, authoring a question over that graph, and abstracting away every directly searchable anchor.

\paragraph{Entity-graph extraction.} Raw pages carry substantial noise that distracts question generation. We distill $G_{\text{sub}}$ into a compact, connected entity graph
\begin{equation}
G_{e}=(V_{e},R_{e})=f_{\text{ext}}(G_{\text{sub}}),\qquad |V_{e}|\le n,
\label{eq:factgraph}
\end{equation}
where $V_{e}$ are salient entities and $R_{e}$ are typed semantic relations that preserve the cross-page link structure of $G_{\text{sub}}$. The extractor keeps only entities and relations that lie on a multi-hop path toward the seed theme, yielding a dense relational skeleton over which questions can be authored precisely.

\paragraph{Multi-hop question generation.} Let $y=\mathrm{theme}(v_{0})$ denote the seed theme, which we take as the target answer. We generate an initial question together with its reasoning path over the entity graph,
\begin{equation}
(q_{0},\,P)=f_{\text{gen}}(G_{e},\,y)\quad\text{s.t.}\quad |P|\ge N \;\wedge\; y\notin q_{0},
\label{eq:qgen}
\end{equation}
where $P=(e_{1}\xrightarrow{\,r_{1}\,}e_{2}\xrightarrow{\,r_{2}\,}\cdots\to y)$ is a path in $G_{e}$ whose traversal uniquely yields $y$. The hard constraint $|P|\ge N$ forces the question to depend on at least $N$ coupled relations.

\paragraph{Anchor abstraction.} Concrete anchors let an agent bypass the intended reasoning by directly searching a surface string. We remove this shortcut with an abstraction operator $\mathcal{A}$ that rewrites every non-answer entity into a descriptive reference,
\begin{equation}
\mathcal{A}(e):\quad \mathrm{name}(e),\ \mathrm{alias}(e)\notin \mathcal{A}(e)\ \wedge\ \mathcal{A}(e)\ \text{uniquely identifies}\ e,
\qquad \forall\, e\in V_{e}\setminus\{y\}.
\label{eq:abs}
\end{equation}
The initial question is then rewritten into its abstracted form
\begin{equation}
\tilde{q}=f_{\text{abs}}(q_{0},\mathcal{A}),
\label{eq:qtilde}
\end{equation}
substituting each mentioned entity $e$ by $\mathcal{A}(e)$ while preserving the reasoning structure and the answer $y$. The resulting $\tilde{q}$ demands disambiguation-by-reasoning rather than string matching.

\subsection{Dual-Criteria Verification}
\label{sec:qa-verify}

We admit a pair $(\tilde{q},y)$ only if it is simultaneously hard and solvable, judged by a reference model $M_{\text{ref}}$ under two settings,
\begin{equation}
c_{\text{diff}}(\tilde{q})=\mathbb{I}\!\left[\,M_{\text{ref}}(\tilde{q})\neq y\,\right],
\qquad
c_{\text{solv}}(\tilde{q})=\mathbb{I}\!\left[\,M_{\text{ref}}(\tilde{q}\mid G_{e})=y\,\right],
\label{eq:criteria}
\end{equation}
and we keep only the intersection
\begin{equation}
\mathcal{D}=\bigl\{\,(\tilde{q},y)\ \bigm|\ c_{\text{diff}}(\tilde{q})\cdot c_{\text{solv}}(\tilde{q})=1\,\bigr\}.
\label{eq:accept}
\end{equation}
The difficulty criterion (closed-book, no tools) discards simple questions the model already answers from memory; the solvability criterion (with $G_{e}$ supplied as context) discards questions whose answer is wrong or non-unique. Answer equality $M_{\text{ref}}(\cdot)=y$ is decided by a semantic-matching judge.
The accepted set $\mathcal{D}$ of abstracted multi-hop questions, each paired with a verified, unique answer, along with some in-house and open-source question sets~\cite{openseeker, lohosearch, redsearcher, xyz}, will be used for trajectory generation.

\section{Training Recipe}
\label{sec:training}

We next train the search agent through iterative cycles of SFT and RL.

\subsection{Supervised Fine-Tuning}
\label{sec:sft}

\paragraph{Trajectory generation.} We prompt a strong teacher $M_{T}$ to solve each question $q\in\mathcal{D}$ under the ReAct paradigm~\cite{react}, interleaving reasoning, tool calls, and observations against live search tools. A trajectory is the resulting sequence of reasoning steps, tool calls, and observations, terminated by a final answer,
\begin{equation}
\tau=\bigl(r_{1},a_{1},o_{1},\,\dots,\,r_{T},a_{T},o_{T},\,r_{T+1},\hat{y}\bigr)
\ \sim\ \pi_{M_{T}}\!\bigl(\cdot \mid q,\ \mathcal{T}\bigr),
\label{eq:traj}
\end{equation}
where $r_{t}$ is the reasoning at step $t$, $a_{t}\in\mathcal{T}$ is a tool call from the tool set $\mathcal{T}=\{\textsc{search},\textsc{scrape}\}$, $o_{t}$ is the returned observation, and $\hat{y}$ is the final answer. Each observation is a document-level summary produced on the fly rather than a raw page, which keeps trajectories within a bounded context budget.

\paragraph{Coarse filtering.} Let $\mathcal{D}_{\text{raw}}=\{(q,\tau)\}$ be the pool of collected trajectories. Every sample must clear a trajectory-level stage that admits $\tau$ only if it is correct, non-degenerate, and non-trivial. Correctness is a two-part gate: the rollout must terminate successfully, and its answer must be judged correct by an LLM judge $J$ against the reference $y^{*}$,
\begin{equation}
c_{\text{corr}}(q,\tau)=\mathbb{I}\!\left[\mathrm{success}(\tau)\right]\cdot\mathbb{I}\!\left[J(q,\hat{y}_{\tau},y^{*})=\textsc{correct}\right].
\label{eq:corr-gate}
\end{equation}
We then remove degenerate trajectories: repetition loops, runaway tool calling, unterminated \texttt{<think>} blocks, and other pathologies. Our primary detector is a sliding-window compression ratio, computed in $O(n)$ over the decoded text,
\begin{equation}
\rho_{\text{cr}}(w)=\frac{|w|}{\bigl|\mathrm{zlib}(w)\bigr|},\qquad
c_{\text{degen}}(\tau)=\mathbb{I}\!\left[\max_{w}\rho_{\text{cr}}(w)\ge \tau_{\text{cr}}\right].
\label{eq:degen}
\end{equation}
Repetitive text compresses far more than fluent text, so any window whose ratio reaches $\tau_{\text{cr}}$ signals a loop, whatever its period and wherever it begins. Auxiliary detectors cover what compression alone can miss: periodically repeating lines, long single-character runs, and consecutive tool calls with byte-identical arguments. Finally we keep only trajectories with at least $K$ tool-call turns, discarding shallow cases a direct lookup could resolve,
\begin{equation}
c_{\text{depth}}(\tau)=\mathbb{I}\!\left[T_{\text{tool}}(\tau)\ge K\right].
\label{eq:depth}
\end{equation}
An exact-duplicate pass over full message sequences then drops byte-identical trajectories while keeping distinct rollouts of the same question, giving the coarse set
\begin{equation}
\mathcal{D}_{\text{sft}}=\operatorname{dedup}\bigl\{(q,\tau)\in\mathcal{D}_{\text{raw}}\ \bigm|\ c_{\text{corr}}\cdot(1-c_{\text{degen}})\cdot c_{\text{depth}}=1\bigr\}.
\label{eq:coarse-set}
\end{equation}

\paragraph{Fine filtering.} Coarse filtering keeps or drops whole trajectories. When a question sits near the capability boundary of the teacher, however, a correct trajectory may still contain a locally poor turn: a redundant search, a hallucinated tool name, or reasoning inconsistent with the action actually taken.
As a turn-level refinement, we label individual turns with an LLM judge. The difficulty is that a turn which looks wasteful in isolation is often a legitimate exploratory step, and because no human is in the loop, the judge has to draw that line on its own.
We therefore induce the criteria from the data rather than hand-crafting them.
Specifically, we sample some trajectories and ask the judge to critique them in free form, then consolidate the recurring failure modes (e.g., misinterpreting the question) into an explicit rubric used in the final judging prompt. For each assistant turn, the judge receives the question, reference answer, and a fixed local window of surrounding turns, and outputs either \textsc{keep} or \textsc{mask}, represented as $m_{t}\in\{0,1\}$. To prevent overly aggressive filtering, we mask at most $10\%$ of the assistant turns in any trajectory. Masked turns remain in the context but are excluded from the training loss, providing cleaner learning signals without discarding useful interaction history.

\paragraph{Training objective.} For each assistant turn, let $C_{<t}$ denote its visible history, reconstructed by the harness's replay operator from the append-only conversation so as to stay byte-identical to what the agent conditioned on at inference. Over $\mathcal{D}_{\text{sft}}$ we maximize the likelihood of the teacher's output,
\begin{equation}
\mathcal{L}_{\text{SFT}}(\theta)=-\,\mathbb{E}_{(q,\tau)\sim\mathcal{D}_{\text{sft}}}
\sum_{t=1}^{T+1} m_{t}\,\log \pi_{\theta}\!\bigl(u_{t}\mid C_{<t}\bigr),
\label{eq:sft-loss}
\end{equation}
where $u_{t}$ is the assistant output at turn $t$, namely the reasoning and tool call $(r_{t},a_{t})$ for $t\le T$ and the final reasoning and answer $(r_{T+1},\hat{y})$ at $t=T+1$. The per-turn mask $m_{t}\in\{0,1\}$ is supplied by fine filtering; observation, user, and system tokens carry zero loss by construction.

\subsection{Reinforcement Learning}
\label{sec:rl}

We then optimize the agent against live search with a group-relative policy gradient. To keep long-horizon rollouts affordable without going fully asynchronous, we place the rollout regime \emph{between} on- and off-policy through request-level partial rollout. To avoid depending on external APIs at training time, we co-locate an in-house model that serves as \emph{both} reward judge and observation summarizer.

\paragraph{Partial rollout and prefix reuse.} Long-horizon rollouts have a heavy tail: a few sessions run far longer than the rest and stall a synchronous step. Rather than discard unfinished work, as task-level streaming does, we interrupt at the request level. Once a step has committed enough completed trajectories, in-flight over-sampled sessions are aborted and resumed at the next step from their committed prefix. Rollouts are organized as a forest whose nodes are message states, each caching its token, loss-mask, log-probability, and weight-version deltas, so a resumed trajectory is a path that splices prefixes generated under different policy weights, a mismatch we correct with truncated importance sampling. Completed turns are therefore reused rather than thrown away, which keeps rollout GPUs busy under a synchronous, co-located schedule at the cost of roughly $2\times$ over-sampling as headroom.


\paragraph{In-house reward and summarization.} Rather than call an external API, we run several FP8 engines of an in-house Qwen3.5-397B-A17B model inside the training cluster, alongside the actor and rollout engines. 
{Acting as a generative reward model (GenRM)}, the model scores a rollout with a binary verdict on the extracted answer $\hat{y}_{\tau}$ against the reference $y^{*}$,
\begin{equation}
R(q,\tau)=\mathbb{I}\!\left[\operatorname{GenRM}\!\bigl(q,\hat{y}_{\tau},y^{*}\bigr)=\textsc{A}\right],
\label{eq:reward}
\end{equation}
with no additive format term, since an empty or mid-thought-truncated answer extracts to $\varnothing$ and already scores $0$ without invoking the judge. As a summarizer, the same engines compress each retrieved page into a short, query-relevant digest that supplies the observation $o_{t}$ of Eq.~\eqref{eq:traj}. This is the only context-reduction mechanism in the rollout, with no message-history pruning or sliding window, so the context the policy conditions on is exactly what is trained on. Serving both roles in-cluster removes the external API dependency from the training loop, and one allocation covers training, rollout, and reward.

\subsection{Iterative Climbing}
\label{sec:climb}

We alternate SFT and RL in iterative cycles. Each round of RL explores the current policy, after which a small set of high-quality rollouts is distilled back into the policy through supervised fine-tuning; RL then resumes from the updated model. We call each such cycle a \emph{climb}. This alternation complements the strengths of the two objectives: RL improves behaviors sampled by the current policy through relative rewards, while supervised fine-tuning directly reinforces rare but successful trajectories instead of relying on their contribution to a group-relative gradient.

For each query $q$, we retain at most one rollout from the existing RL samples. Let $\bar{R}(q)$ denote the pass rate within its rollout group. We select queries with $0<\bar{R}(q)\le 1/2$, targeting cases that are solvable but not yet reliable. Among successful rollouts, we require at least $K_{\text{rft}}$ tool-call turns and then select the shortest valid trajectory:
\begin{equation}
\mathcal{S}_{q}
=
\bigl\{
\tau_i
\mid
R(q,\tau_i)=1,\;
T_{\text{tool}}(\tau_i)\ge K_{\text{rft}}
\bigr\},
\qquad
\tau^{\star}_{q}
=
\operatorname*{arg\,min}_{\tau\in\mathcal{S}_{q}}
T_{\text{tool}}(\tau).
\end{equation}
The depth constraint filters out lucky or trivial solutions, while the shortest-trajectory criterion discourages unnecessary search. The resulting set is deduplicated and used for SFT before the next RL round. Since the difficulty band is defined by the current policy's pass rate, it automatically shifts toward harder examples as the policy improves, providing a simple self-paced curriculum and a natural stopping signal as the candidate pool diminishes.
Further details will be released in the future.

\section{Evaluation}
\label{sec:exp}

\subsection{Setup}
\label{sec:exp-setup}

\paragraph{Benchmarks.}
We evaluate our models on four challenging benchmarks: BrowseComp~\cite{browsecomp}, BrowseComp-ZH~\cite{browsecompzh}, DeepSearchQA~\cite{deepsearchqa}, and {HLE}~\cite{hle}. BrowseComp evaluates an agent's ability to identify long-tail entities and provide concise answers based on multiple indirect and mutually constraining clues. BrowseComp-ZH follows the same evaluation setting but focuses on Chinese-language sources. DeepSearchQA evaluates the comprehensiveness of search-based answers rather than the correctness of a single answer span, with scores reflecting the proportion of required evidence successfully recovered by the agent. HLE evaluates expert-level academic reasoning across a broad range of disciplines, where information retrieval is expected to complement rather than substitute for domain knowledge. We evaluate the text-only subset of HLE.

\paragraph{Implementation Details.}
{We initialize Iris-mini and Iris-pro from Qwen3.6-35B-A3B and Qwen3.5-397B-A17B respectively}, both of which adopt a {mixture-of-experts (MoE)} architecture with a 256K-token context window. For SFT, the models are trained for two epochs with a global batch size of 64 and a maximum sequence length of $262{,}144$ tokens. For RL, we use the open-source Relax framework~{\cite{relax}} as the underlying training engine.

To prevent models from exploiting benchmark leakage, we block access to the Hugging Face dataset and Space pages that host the benchmark questions and answers (i.e., \texttt{huggingface.co/datasets} and \texttt{huggingface.co/spaces}), enforced at three points: removed from search results, refused on scrape, and caught by a post-hoc guard in the tool manager even when the model supplies the URL from memory.

\paragraph{Context Management.}
We believe that specifically designing different {CM} strategies for different benchmarks is of little practical significance. Therefore, for all benchmarks, our main results are reported under two settings: {without CM}, and with the \emph{discard-all} strategy of DeepSeek-V3.2~\cite{deepseek}. In Section~\ref{sec:exp-cm}, we further compare a broader range of {CM} strategies and evaluate their performance across different benchmarks.

\paragraph{Evaluation Protocol.}
We perform a single rollout (pass@1) for each question. The resulting final answer is evaluated against the corresponding reference answer using an LLM-based judge with each benchmark's official evaluation prompt. DeepSearchQA is evaluated using F1, whereas BrowseComp, BrowseComp-ZH, and HLE are evaluated using accuracy. All benchmarks are evaluated under a consistent configuration, including the same tool set, context-length limit, and maximum turn budget.

\subsection{Experimental Results}
\label{sec:exp-main}

Table~\ref{tab:main-results} summarizes the main results on the four benchmarks. For our models we report the {discard-all} setting, which we treat as the default configuration; {the remaining strategies are examined in Section~\ref{sec:exp-cm}}.

In the 30--35B parameter range, {Iris-mini} achieves the best performance on three benchmarks: BrowseComp (82.2), BrowseComp-ZH (84.8), and HLE (52.3). On BrowseComp, it outperforms the strongest model in the same parameter range, XYZ-Aquila-mini~{\cite{xyz}}, by 3.4 points (82.2 vs. 78.8). On DeepSearchQA, {it} achieves an F1 score of 86.9, which remains below XYZ-Aquila-mini (89.5).

In the $\sim$400B parameter range, {Iris-pro} leads or matches the best result on all four benchmarks. It achieves the highest score on BrowseComp (88.6), outperforming XYZ-Aquila-pro by 3.8 points, the highest F1 on DeepSearchQA (92.9), and the highest accuracy on HLE (56.4), ahead of XYZ-Aquila-pro by 3.1 points. On BrowseComp-ZH it ties XYZ-Aquila-pro at 85.1.

These results demonstrate that our systems are competitive relative to models with substantially larger parameter counts. 
{Iris-mini} approaches the performance of 1T-scale models such as Kimi-K2.6 and DeepSeek-V4-Pro on BrowseComp. This highlights the effectiveness of our training approach in specializing a relatively compact model for web search and long-horizon information-seeking tasks.
{Iris-pro} further demonstrates strong intrinsic search capabilities: even in the standard ReAct setting, it achieves performance comparable to or better than MiroThinker-H1 and Apodex-1.0-H under their heavy-compute configurations. These results reflect substantial improvements in the underlying search intelligence of our models.
Nevertheless, our models still leave a gap relative to the most capable frontier systems. Closing this gap remains an important direction for future work.



\begin{table}[!t]
\centering\small
\setlength{\tabcolsep}{10pt}
\renewcommand{\arraystretch}{1.4}
\caption{Main results on the four search benchmarks with {CM} enabled for each system.
For our models we report the {discard-all} setting throughout; the remaining strategies are analyzed in Table~\ref{tab:cm-ablation}.
HLE uses the text-only setting by default; results evaluated on the full set are marked with $^{f}$.
$^{r}$ indicates that the corresponding result was reproduced by the XYZ-Aquila Team.}
\label{tab:main-results}
\vspace{6pt}
\setlength{\aboverulesep}{0pt}\setlength{\belowrulesep}{0pt}
\begin{tabular}{l c cccc}
\toprule
\textbf{Model} & \textbf{Size} & \textbf{BrowseComp} & \textbf{BrowseComp-ZH} & \textbf{DeepSearchQA} & \textbf{HLE} \\
\midrule
\multicolumn{6}{l}{\textit{30$\sim$35B}} \\
MiroThinker-1.7-mini & 30B & 67.9 & 72.3 & -- & 36.4 \\
FORT-Searcher        & 30B & 72.2 & 75.0 & -- & -- \\
Agents-A1            & 35B & 75.5 & -- & -- & 47.6 \\
Nex-N2-mini          & 35B & 74.1 & 79.6$^{r}$ & 87.2$^{r}$ & 37.1$^{r}$ \\
Apodex-1.0-mini      & 35B & 71.5 & 80.6 & 82.2 & 46.8 \\
XYZ-Aquila-mini      & 35B & 78.8 & 82.9 & \textbf{89.5} & 51.1 \\
{\textbf{Iris-mini}} & 35B & \textbf{82.2} & \textbf{84.8} & 86.9 & \textbf{52.3} \\
\midrule
\multicolumn{6}{l}{\textit{$\sim$ 400B}} \\
MiroThinker-1.7      & 397B & 74.0 & 75.3 & -- & 42.9 \\
Nex-N2-Pro           & 397B & 83.7 & 79.6$^{r}$ & 92.3$^{r}$ & 50.0$^{r}$ \\
Apodex-1.0           & 397B & 75.5 & 82.6 & 84.6 & 49.0 \\
XYZ-Aquila-pro       & 397B & 84.8 & \textbf{85.1} & 92.5 & 53.3 \\
{\textbf{Iris-pro}} & 397B & \textbf{88.6} & \textbf{85.1} & \textbf{92.9} & \textbf{56.4} \\
\midrule
\multicolumn{6}{l}{\textit{Frontier / heavy-compute}} \\
MiroThinker-H1       & --   & 88.2 & \textbf{84.4} & -- & 47.7 \\
Apodex-1.0-H         & --   & 90.3 & 84.1 & 94.4 & 60.8 \\
Kimi-K2.6            & 1T   & 83.2 & -- & 92.5 & 54.0$^{f}$ \\
GLM-5.2              & 744B & -- & -- & -- & 54.7 \\
DeepSeek-V4-Pro (Preview) & 1.6T & 83.4 & -- & -- & 48.2 \\
Claude Fable 5       & --   & 88.0 & -- & 94.2 & \textbf{64.5}$^{f}$ \\
GPT-5.6 Sol          & --   & 90.4 & -- & -- & 58.0$^{f}$ \\
Kimi-K3              & 2.8T & \textbf{91.2} & -- & \textbf{95.0} & 56.0$^{f}$ \\
Qwen3.8-Max          & 2.4T & -- & -- & -- & 56.2$^{f}$ \\
\bottomrule
\end{tabular}
\end{table}

\subsection{Analysis}
\label{sec:exp-cm}

Table~\ref{tab:cm-ablation} investigates the effect of {CM} on our search agents. In addition to \emph{discard-all}, which clears the entire interaction history and restarts the question from scratch when the running context reaches a predefined threshold before a final answer is obtained, we also evaluate the \emph{retry} strategy proposed by MiroThinker~\cite{mirothinker}.
When an attempt fails to produce a valid answer, \emph{retry} summarizes the failed attempt into a compact description of the unsuccessful search experience and appends it to the task for the next attempt. In this sense, retry can be viewed as a form of {discard-all} with progressively accumulated prior knowledge about what has already been explored and ruled out. We do not consider additional inference-time verification or re-answering procedures adopted by some systems, such as the \emph{reverify} and \emph{reanswer} strategies used by XYZ-Aquila~{\cite{xyz}} on BrowseComp-ZH and related benchmarks, as these introduce additional test-time mechanisms beyond {CM} itself.

\begin{table}[!t]
\centering\small
\setlength{\tabcolsep}{8pt}
\renewcommand{\arraystretch}{1.25}
\caption{Effect of {CM}. For each system, w/o denotes the baseline without {CM}. The subsequent rows show the results of different {CM} strategies, with the change relative to the corresponding baseline reported in parentheses.}
\label{tab:cm-ablation}
\vspace{6pt}
\setlength{\aboverulesep}{0pt}\setlength{\belowrulesep}{0pt}
\begin{tabular}{l cccc}
\toprule
\textbf{Context management} & \textbf{BrowseComp} & \textbf{BrowseComp-ZH} & \textbf{DeepSearchQA} & \textbf{HLE} \\
\midrule
\multicolumn{5}{l}{\textit{OpenSeeker-v2 (30B)}} \\
w/o                 & 46.0 & 58.1 & -- & 34.6 \\
\midrule
\multicolumn{5}{l}{\textit{REDSearcher (30B)}} \\
w/o                 & 42.1 & 49.8 & -- & 34.3 \\
discard-all         & {57.4} (+15.3) & {58.2} (+8.4) & -- & -- \\
\midrule
\multicolumn{5}{l}{\textit{FORT-Searcher (30B)}} \\
w/o                 & 55.9 & 62.1 & -- & -- \\
discard-all         & {72.2} (+16.3) & {75.0} (+12.9) & -- & -- \\
\midrule
\rowcolor{gray!15} \multicolumn{5}{l}{{\textit{Iris-mini (35B)}}} \\
\rowcolor{gray!15} w/o                 & 64.7 & 72.3 & 81.0 & 43.2 \\
\rowcolor{gray!15} retry               & -- & 83.0 (+10.7) & 89.1 (+8.1) & 52.0 (+8.8) \\
\rowcolor{gray!15} discard-all         & 82.2 (+17.5) & 84.8 (+12.5) & 86.9 (+5.9) & 52.3 (+9.1) \\
\rowcolor{gray!15} discard-all + retry & \textbf{85.9} {(+21.2)} & \textbf{85.1} (+12.8) & \textbf{89.9} (+8.9) & \textbf{52.4} (+9.2) \\
\midrule
\rowcolor{gray!15} \multicolumn{5}{l}{{\textit{Iris-pro (397B)}}} \\
\rowcolor{gray!15} w/o                 & 72.6 & 76.8 & 86.4 & 50.8 \\
\rowcolor{gray!15} retry               & -- & 84.1 (+7.3) & 92.3 (+5.9) & \textbf{56.6} (+5.8) \\
\rowcolor{gray!15} discard-all         & 88.6 (+16.0) & \textbf{85.1} (+8.3) & 92.9 (+6.5) & {56.4 (+5.6)} \\
\rowcolor{gray!15} discard-all + retry & \textbf{90.3} (+17.7) & \textbf{85.1} (+8.3) & \textbf{93.4} (+7.0) & \textbf{56.6} (+5.8) \\
\bottomrule
\end{tabular}
\end{table}

Several prior works explicitly report results without any {CM}, including OpenSeeker-v2~\cite{openseekerv2}, REDSearcher~\cite{redsearcher}, and FORT-Searcher~\cite{fortsearcher}. We believe such results are particularly informative because they provide a more direct measurement of the search agent itself. Without an additional {CM} mechanism, the reported performance is determined primarily by the quality of the training data and the search behavior learned by the model, rather than by how much performance can be recovered through an external inference-time wrapper.
We therefore also report these {no-CM} results in Table~\ref{tab:cm-ablation} for a more direct comparison. This distinction is important because many existing results are reported only with {CM} enabled, making it difficult to disentangle the intrinsic capability of the agent from the gains introduced by the inference-time strategy.

Our models demonstrate a clear advantage in both regimes. Without any {CM}, {Iris-mini} achieves 64.7 on BrowseComp and 72.3 on BrowseComp-ZH, outperforming FORT-Searcher (55.9 and 62.1), OpenSeeker-v2 (46.0 and 58.1), and REDSearcher (42.1 and 49.8). {Iris-pro raises these two scores by a further} 7.9 and 4.5 points, respectively. More importantly, the advantage is retained after {CM} is introduced, indicating that our gains are not solely attributable to the inference-time context strategy. Instead, the underlying search capability learned by our models already provides a strong foundation, which {CM} can further exploit.

Across the evaluated settings, {CM} consistently improves performance over the corresponding {no-CM} baseline, with substantially larger gains observed for {Iris-mini} than for {Iris-pro}. What differs is not the size of the budget but how quickly each model consumes it: the smaller model needs more steps to resolve the same set of constraints, so it reaches the limit more often and CM has correspondingly more to recover. CM is therefore most valuable when a model has already learned effective search behaviors but exhausts its context before it can bring them to bear.

The gains are also strongly benchmark-dependent. For Iris-mini, CM is worth up to {$21.2$} points on BrowseComp, against at most $12.8$ on BrowseComp-ZH, $9.2$ on HLE, and $8.9$ on DeepSearchQA. 
This ordering is not explained by how much room each benchmark leaves: HLE has the lowest no-context baseline of the four at $43.2$, yet it gains less than BrowseComp.
What the ordering does follow is how often a session actually runs out of context. BrowseComp-style tasks require long-horizon information seeking, in which the agent repeatedly retrieves, filters, and integrates evidence across many search steps; as the interaction history grows, the context itself becomes the binding constraint, and resetting the accumulated tool history buys additional search on the same question. 
HLE instead places its emphasis on expert-level knowledge and reasoning, for which web retrieval is complementary rather than the primary source of the answer. Its deficit is therefore not a context problem, and extending the effective search horizon recovers correspondingly less.


Another notable observation is how tightly the two model scales converge on BrowseComp-ZH. Three configurations land on exactly the same score of $85.1$: {Iris-mini} under discard-all with retry, and {Iris-pro} under both discard-all and discard-all with retry. On a benchmark of $289$ questions this corresponds to $246$ correct answers in all three cases, suggesting that, at least on this benchmark, the remaining performance is constrained by factors beyond model capacity. We discuss the quality and characteristics of BrowseComp-ZH {in Appendix~\ref{sec:case}}.

Finally, combining {discard-all} with retry achieves the strongest result in most settings and pushes {Iris-pro} above 90 on BrowseComp. However, this improvement comes at a substantial inference cost, since each retry requires another full search attempt. We therefore regard retry as an interesting exploration of the upper bound achievable through increasingly aggressive inference-time strategies, rather than as the primary configuration for reporting model performance. Accordingly, Table~\ref{tab:main-results} reports the {discard-all} results for all models, even when retry can produce a higher score. This concern is not ours alone: the same point is made in~\cite{deepseek}, which argues that actual compute cost has to be accounted for when benchmarking {CM} strategies. Our goal is not simply to maximize a benchmark score through increasingly elaborate inference-time procedures, but to train a genuinely capable search agent. Benchmark performance captures only part of an agent's overall capability, and we therefore consider evaluations in a broader range of realistic search scenarios to be an important direction beyond these benchmark results.

\section{Conclusion: Beyond Search}
\label{sec:conclusion}

In this report, we presented an end-to-end recipe for building search agents, covering data construction, training, and evaluation, and instantiated it at two scales: Iris-mini and Iris-pro. Training questions are reverse-constructed from the hyperlink structure of a web corpus, stripped of directly searchable anchors, and retained only when a reference model fails them in the closed-book setting but solves them once the supporting evidence is provided. A strong teacher then turns these questions into trajectories, which are filtered at both the trajectory and turn levels before supervised training. The resulting policy is subsequently optimized against live search, with the judge and observation summarizer served within the training cluster; the two stages are alternated so that each round of exploration is consolidated before the next begins.
As an initial exploration by our team, this report focuses on several widely recognized benchmarks in the search community, where our models achieve the best overall performance within their respective parameter ranges. This evaluation provides a useful first view of the capabilities that can be developed through our search-agent training recipe, while leaving a much broader space of agentic scenarios to explore. 

One observation from our experiments points beyond search itself. Both the synthesized search data and the search-specialized models, the latter used as teachers for on-policy distillation, transferred positively to domains that were not explicitly targeted, including General Tool Use (BFCL and $\tau$-bench) and Cowork (OfficeQA and APEX). We interpret this as evidence that search may be better viewed as an atomic capability than as a vertical specialization: the behaviors it induces are not limited to web browsing and appear reusable wherever an agent must act under incomplete information. If this observation generalizes, search data and search-derived teachers may be useful throughout training rather than confined to a separate specialization stage, with their value measured not only by browsing performance but also by broader agentic competence. This is the direction we intend to pursue.

\section{Contributions}
\label{sec:contributions}

\noindent
{Authors are listed in order of contribution.}

\vspace{4pt}
\begingroup
\renewcommand{\thefootnote}{\fnsymbol{footnote}}
\noindent
{%
Ziyuan Liu\footnotemark[1],
Hengqi Liu\footnotemark[1],
Zichuan Wang\footnotemark[1],
Yang Qin\footnotemark[1]\footnotemark[2]\footnotemark[3],
Jiachen Liang,
Xu Chu,
Shaowei Chen,
Yuantao Gu,
Zhaokai Luo,
Yao Hu,
Mu Chuan\footnotemark[3]}
\footnotetext[1]{{Equal contribution.}}
\footnotetext[2]{{Project lead.}}
\footnotetext[3]{{Corresponding author.}}
\endgroup

\bibliographystyle{styles/colm2026_conference}
\bibliography{bibliography/references}

\appendix

\section{{A Case of Ground-Truth Inconsistency}}
\label{sec:case}

\begin{figure}[h]
\centering
\includegraphics[width=0.8\linewidth]{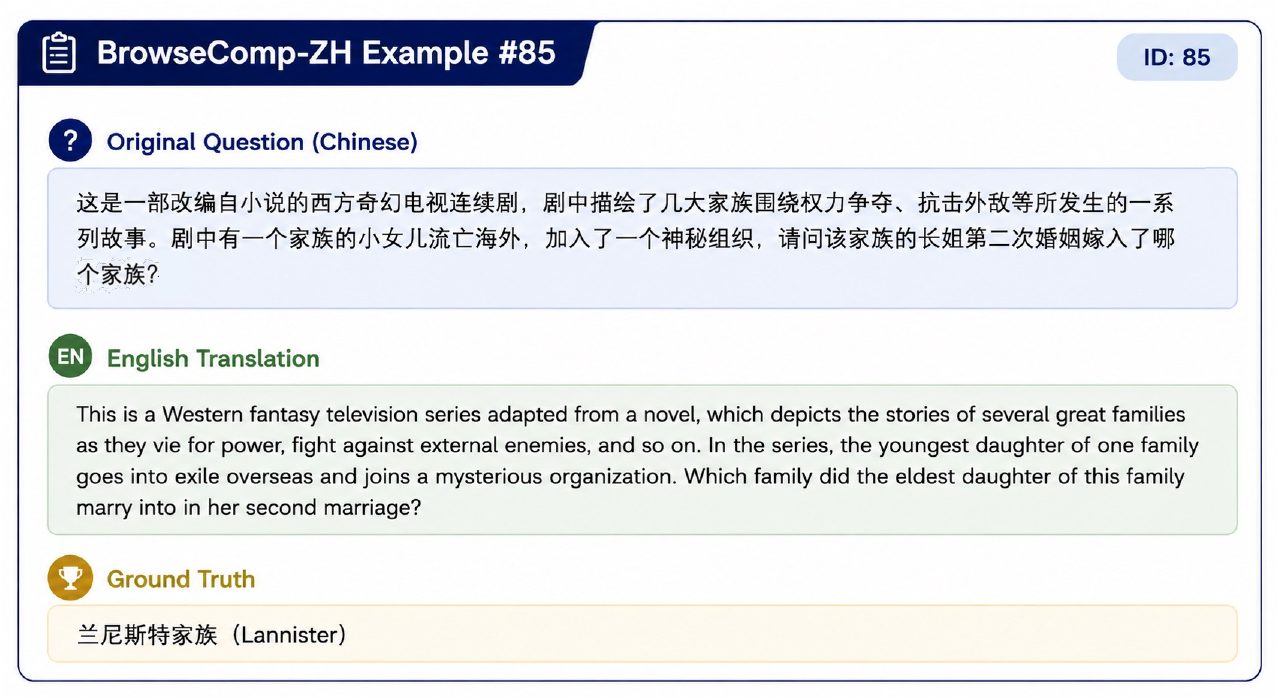}
\caption{An illustrative case from BrowseComp-ZH (Question 85), where our search agent returns the answer ``Bolton'', while the official ground truth is ``Lannister''.}
\label{fig:lannister}
\end{figure}

An interesting case is shown in Figure~\ref{fig:lannister}. In Question 85 of BrowseComp-ZH, the official ground truth is \emph{Lannister}, whereas our search agent returns \emph{Bolton}. The clues in the question identify the target series as \emph{Game of Thrones}: the youngest daughter who goes into exile is Arya Stark, who joins the Faceless Men, identifying the family as House Stark and its eldest daughter as Sansa Stark.
The key is to trace Sansa's two formal marriages. Although she is initially betrothed to Joffrey Baratheon, they \textbf{never} marry. Her first formal marriage is to Tyrion Lannister in Season 3, arranged by Tywin Lannister. She later marries Ramsay Bolton in Season 5, when Petyr Baelish arranges the match. Thus, Sansa's second marriage is to \emph{House Bolton}, not \emph{House Lannister}. As a result, our system did not receive credit for this question under the official evaluation, despite the answer being directly supported by the events in the series.
This case points to a potential inconsistency between the benchmark annotation and the underlying source material. Such observations further motivate us to develop new search benchmarks with higher-quality annotations and broader, more reliable capability coverage.

\end{document}